\documentclass[10pt,twocolumn,letterpaper]{article}

\usepackage{wacv}
\usepackage{times}
\usepackage{epsfig}
\usepackage{graphicx}
\usepackage{amsmath}
\usepackage{amssymb}

\usepackage{import}
\usepackage{epstopdf}
\usepackage{caption}
\usepackage{subcaption}
\usepackage{tabu, tabularx}
\usepackage{verbatim}
\usepackage{nicefrac}
\usepackage{color}

\DeclareMathOperator*{\argmin}{argmin}

\wacvfinalcopy % *** Uncomment this line for the final submission

\def\wacvPaperID{203} % *** Enter the wacv Paper ID here

\ifwacvfinal\fi
\begin{document}

%%%%%%%%% TITLE
\title{A Temporal Sequence Learning for Action Recognition and Prediction}

% Authors at the same institution
%\author{First Author \hspace{2cm} Second Author \\
% Authors at different institutions
\author{Sangwoo Cho, Hassan Foroosh \\
University of Central Florida \\
{\tt\small swcho@knights.ucf.edu, foroosh@cs.ucf.edu}
%\and
%Second Author \\
}

\maketitle
\ifwacvfinal\thispagestyle{empty}\fi

%%%%%%%%% ABSTRACT
\begin{abstract}
	
	In this work\footnote {This work was supported in part by the National Science Foundation under grant IIS-1212948.}, we present a method to represent a video with a sequence of words, and learn the temporal sequencing of such words as the key information for predicting and recognizing human actions. We leverage core concepts from the Natural Language Processing (NLP) literature used in sentence classification to solve the problems of action prediction and action recognition. Each frame is converted into a word that is represented as a vector using the Bag of Visual Words (BoW) encoding method. The words are then combined into a sentence to represent the video, as a sentence. The sequence of words in different actions are learned with a simple but effective Temporal Convolutional Neural Network (T-CNN) that captures the temporal sequencing of information in a video sentence. We demonstrate that a key characteristic of the proposed method is its low-latency, i.e. its ability to predict an action accurately with a partial sequence (sentence). Experiments on two datasets, \textit{UCF101} and \textit{HMDB51} show that the method on average reaches 95\% of its accuracy within half the video frames. Results, also demonstrate that our method achieves compatible state-of-the-art performance in action recognition (i.e. at the completion of the sentence) in addition to action prediction.

\end{abstract}

%%%%%%%%% BODY TEXT
\section{Introduction} \label{introduction}

Video-based action recognition is an active research area due to its important practical applications in many areas, such as video surveillance, behavior analysis, and human-computer interaction. The action recognition task is accomplished after acquiring the entire video, while action prediction is different in the sense that it aims at classifying the action with shortest possible latency, i.e. classify as early as possible as the frames come in. The capability of predicting an action early is crucial in both surveillance systems and human-computer interaction. The two tasks of action prediction and recognition have often been researched separately under different settings and constraints.

\begin{figure}[!htb]
	\begin{center}
		\includegraphics[width=1.0\linewidth]{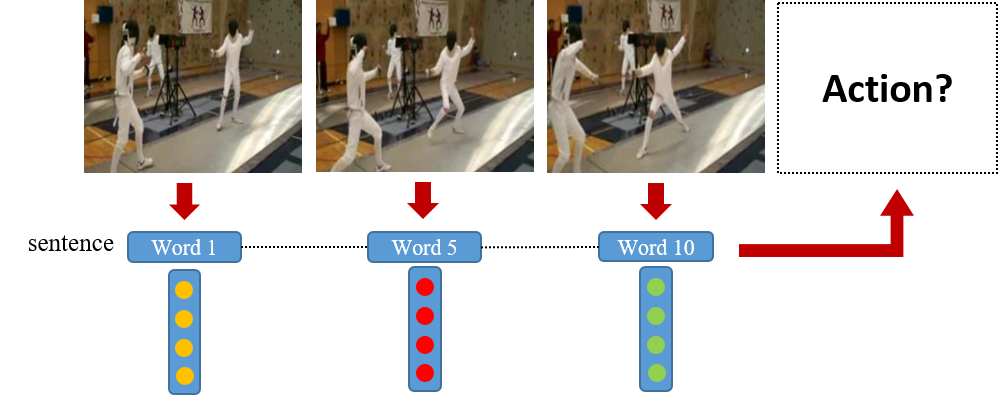}
	\end{center}
	\vspace*{-0.5cm}
	\caption{Given a partial or a full video frames, our goal is to classify the correct action. Each frame is converted to a corresponding word and the sequence of words is trained to predict an activity.}
	\label{fig:overall_desc}
	\vspace*{-0.5cm}
\end{figure}

A video contains two important pieces of information: appearances and motions. These information are complementary, and therefore an accurate prediction relies on the ability to extract the information with low latency, i.e. as early as possible in the temporal sequence. However, extracting effective information (whether for prediction or recognition) is non-trivial due to a number of difficulties such as viewpoint changes, camera motions, and scale variations, to name a few. It is thus crucial to design an effective and generalized representation of a video. Convolutaional Neural Networks (ConvNets)~\cite{intro_ref1} have been playing a key role in solving hard problems in various areas of computer vision, e.g. image classification~\cite{intro_ref1, related_ref1, intro_ref2} and human face recognition~\cite{intro_ref4}. ConvNets also have been employed to solve the problem of action recognition~\cite{ar_ref1, ar_ref12, ar_ref13, ar_ref14} in recent literature. 

Data-driven supervised learning enables to achieve discriminating power and proper representation of a video from raw data. However, ConvNets for action recognition have not shown a significant performance gain over the methods utilizing hand-crafted features~\cite{feat_ref3, related_ref7}. We speculate that the main reason for the lack of big impact is that ConvNets employed in action recognition do not take full advantage of temporal sequencing or order. Recently some methods~\cite{intro_ref5, ar_ref8} attempted to capture long-term temporal information. However, they require excessive computation for a long video. 

In this work, inspired by key ideas from NLP, we represent each frame as a word and a video as a sequence of words. The sequence of words, or a sentence, is a new video representation as shown in Fig.~\ref{fig:overall_desc}. We call the word as \textit{ActionWord}. We use the standard BoW~\cite{bovw_ref8} framework to encode each visual feature as an assigned word in a codebook. The sequence of words then is learned with a simple but effective CNN architecture capturing the sequential order of temporal information. This method is flexible to input size, and hence is applicable to any length of videos. The capability to adopt a variable-size input, combined with low latency versus high accuracy makes the method particularly powerful for both action prediction and action recognition.

Our key contributions can thus be summarized as follows: (i) A new representation for video data as a sequence of words that inherently captures temporal order and sequencing of information. (ii) An effective ConvNet that learns such temporal sequencing to predict with low latency an action. (iii) The ability of the method to maintain state-of-the-art accuracy in both predition and recognition with the challenging datasets, such as UCF101 and HMDB51. (iv) The entire system is easy to implement and is trained with a small amount of computational cost compared to other methods employing ConvNets.

%------------------------------------------------------------------------
\section{Related Work} \label{relatedwork}

Several works using ConvNets to acquire temporal information for action recognition have been studied. In ~\cite{ar_ref7}, hand crafted features are used in the pooling layer of ConvNet to take advantage of both merits of hand-designed and deep learned features. Temporal information from optical flow is explicitly learned with ConvNets in ~\cite{ar_ref1} and the result is fused with the effect of the trained spatial (appearance) ConvNet. \cite{ar_ref5} merges the ConvNet architecture of the two streams ConvNets~\cite{ar_ref1} to capture spatio-temporal information. Although the aforementioned approaches capture temporal information in small time windows, they fail to capture long-range temporal sequencing information that contains long-range ordered information. 

Several works modeling a video-level representation or modeling long temporal information with ConvNets have also been investigated. \cite{ar_ref6} proposes a method that employs a ranking function to generate a video-wide representation that captures global temporal information. In ~\cite{related_ref8}, a HMM model is used to capture the appearance transitions and a max-margin method is employed for temporal information modeling in a video. \cite{ar_ref8, related_ref8, related_ref9} utilize LSTM~\cite{related_ref1} unit in their ConvNets and attempt to capture long-range temporal information. However, the most natural way of representing a video as long-range ordered temporal information is not fully exploited.

Action prediction is to recognize an action with a partial amount of video data. The task may be considered as a subset of the action recognition problem, in a sense that the input data is limited. \cite{ap_ref2} proposes the integral BoW and dynamic BoW to model an action in a particular stage. Sparse coding is used to compute activity likelihood of video segments~\cite{ap_ref1}. A max-margin learning method for prediction is proposed in \cite{ap_ref1}, where human activity is represented in a hierarchical way. \cite{ap_ref4, ap_ref6} employ structured SVM to detect an event and capture global and local dynamics of motions. However, the performance of the above methods are not comparable to our results and they are not applicable to large-scale datasets, such as UCF101~\cite{ar_ref4}.

Our work is inspired by a key idea of sentence classification ~\cite{text_ref1, text_ref3, text_ref4, text_ref5} in NLP. We convert from the domain of images to a domain of words to represent each frame as a word and hence represent a video as a sequence of words, i.e. a sentence. In NLP, words in a sentence are often represented in the form of vectors, see for instance word2vec~\cite{related_ref2} and Glove~\cite{related_ref3}. In order to acquire a similar frame-level representation, we adopted the standard BoW ~\cite{bovw_ref8} encoding method to handle large variability of motions and appearances in video data. It is worth noting, however, that our method can adopt any type of frame-level features to represent video frames as words. 

Various ConvNet arichitectures~\cite{text_ref1, text_ref3, text_ref4, text_ref5} have been taken into account for sentence classification. \cite{text_ref4} utilizes dynamic pooling ConvNets for modeling sentences. In ~\cite{text_ref3, text_ref5}, a simple 1D ConvNet is employed to classify sentences, and LSTM unit is additionally inserted in ~\cite{text_ref1}. 
Similarly, we utilize a simple but effective ConvNet for learning video word sequencing for action prediction and recognition applicable to large-scale datasets.

%------------------------------------------------------------------------
\section{Approach} \label{approach}

\begin{figure*}[!t]
	\begin{center}
		%\fbox{\rule{0pt}{2in} \rule{0.8\linewidth}{0pt}}
		\includegraphics[width=0.9\linewidth]{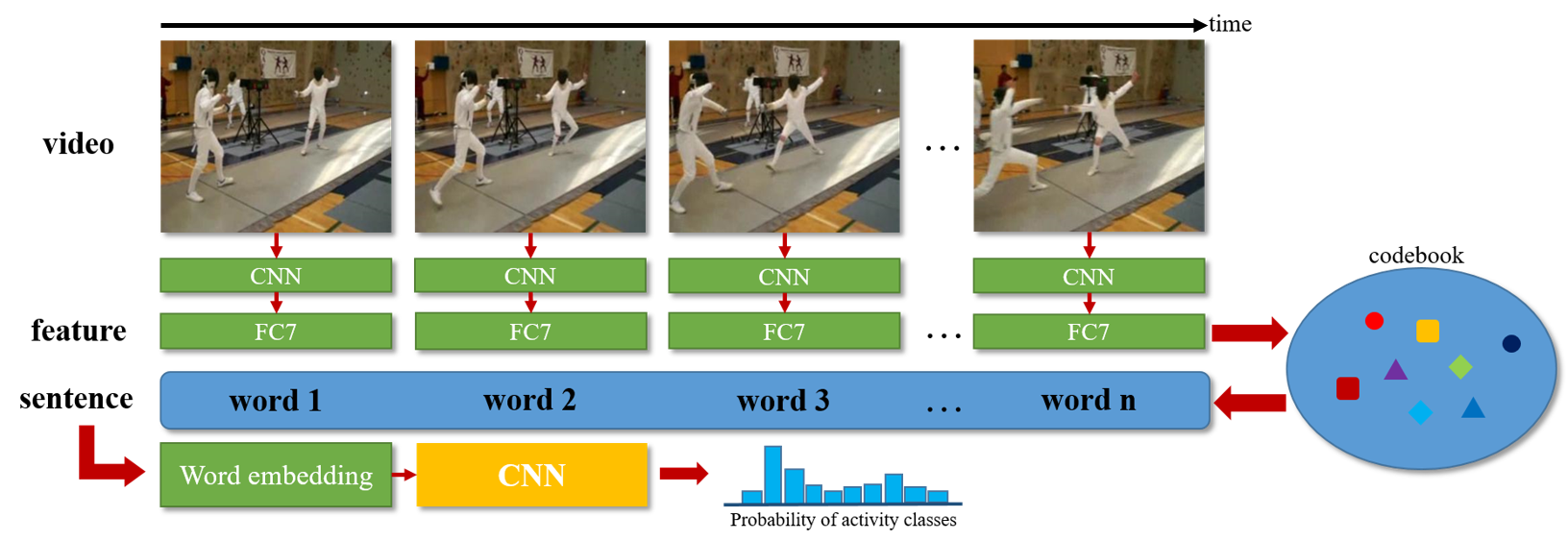}
	\end{center}
	\vspace*{-0.5cm}
	\caption{Pipeline of our method for action prediction/recognition. First, we extract features from video frames using a trained CNN. We then generate a codebook to assign each feature as \textit{ActionWord} as explained in section ~\ref{apporach:bovw}. Finally, a sequence of \textit{ActionWord}s is learned with a sequence learning CNN to classify actions, as described in section ~\ref{apporach:learn_video}.}
	\label{fig:detail_desc}
\end{figure*}

In this section, we give a detailed description of the proposed word encoding and word sequence learning. The pipeline of our method is illustrated in Fig.~\ref{fig:detail_desc}.

%-------------------------------------------------------------------------
\subsection{BoW Framework for Word Representation} \label{apporach:bovw}
\noindent \textbf{Feature Extraction}: Since the approaches based on ConvNets~\cite{ar_ref1, ar_ref2, ar_ref5, ar_ref7} recently have achieved competitive results, we utilize \textit{deep-learned} features. In~\cite{ar_ref1}, a two-stream ConvNet is trained with stacked optical flows and frames. We follow the two-stream ConvNet method and extract $\mathrm{N}$ features $\{\mathrm{x}_1, \cdots ,\mathrm{x}_{\mathrm{N}}\}$, where $\mathrm{x}_t \in \mathbb{R}^D$, every $\mathrm{T}$ frame from all videos using the two trained networks. The extracted features are the output vectors of fully connected (FC) layers on both ConvNets and the dimension is $D$.
Note that the input frames of consecutive temporal features are overlapped by $(\mathrm{L}-\mathrm{T})$ frames, when $\mathrm{L}>\mathrm{T}$, as we train the temporal network with $\mathrm{L}$ stacked frames. The temporal ConvNet is trained with $\mathrm{L}=10$ and $\mathrm{T}$ is set to 5 to consider partial overlap between consecutive temporal features. 
Also, it should be noted that any frame-wise feature extraction techniques can be utilized to represent each frame as a vector.

\noindent \textbf{Codebook Generation}: A codebook is generated to represent each feature as an \textit{ActionWord}. A typical choice for constructing the codebook is $k$-means~\cite{bovw_ref1} or Gaussian Mixture Model (GMM)~\cite{bovw_ref1}.  In our method, we used the method of approximate $k$-means~\cite{bovw_ref2} to construct the codebook with all extracted features from training videos. 
The generated $K$ clusters $\{\mathrm{c}_1, \cdots ,\mathrm{c}_K\}$, where $\mathrm{c}_k \in \mathbb{R}^D$, are employed to both training and testing videos.

\begin{figure}[!tbp]
	\centering
	\begin{subfigure}[b]{.26\textwidth}
		\centering
		\includegraphics[width=0.9\linewidth]{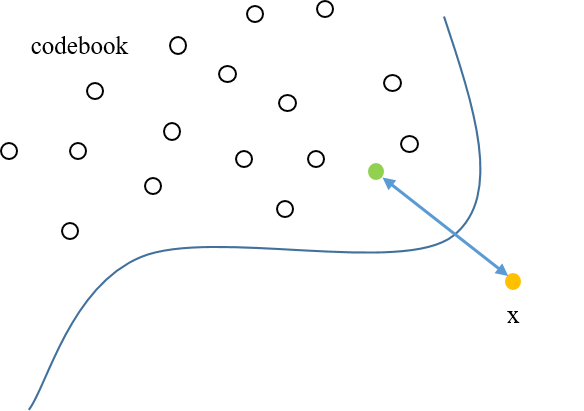}
		\caption{Hard Assignment}
		\label{fig:vq}
		\vspace*{-0.3cm}
	\end{subfigure}%
	\begin{subfigure}[b]{.26\textwidth}
		\centering
		\includegraphics[width=0.9\linewidth]{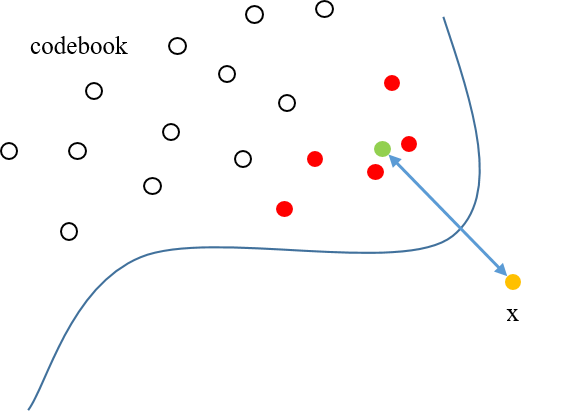}
		\caption{$k$-Soft Assignment}
		\label{fig:5nn_vq}
		\vspace*{-0.3cm}
	\end{subfigure}
	\caption{Feature encoding methods}
	\label{fig:encoding_vote}
	\vspace*{-0.4cm}
\end{figure}

\noindent \textbf{Codeword Assignment}: For coding a video, every extracted video frame feature vector $\mathrm{x}$ needs to be mapped to one of the vectors in the codebook, i.e. to one \textit{ActionWord} that best represents the frame-level visual information at time $\mathrm{T}$. We consider two voting based assignment methods: Hard assignment (HA)~\cite{bovw_ref6} (or Vector Quantization) and soft assignment (SA)~\cite{bovw_ref7}, and a direct assignment as described below.

\textbf{- Hard Assignment}: With HA, \textit{ActionWord} $A$, is simply associated with its nearest codeword to the feature as shown in Fig.~\ref{fig:vq}. The nearest codeword is determined as the one best correlated with the feature vector $\mathrm{x_n}$. The assigned word number (label) for each feature is a sequential number from 1 to $K$.
%Each \textit{ActionWord} has a corresponding vector $\mathrm{c}$ that is one of the $K$ cluster centers.
%When constructing the codebook, $k$-nearest codewords are computed for each feature $\mathrm{x}$.
\begin{equation}
A_{HA_i} = \argmin_i \left\|\mathrm{x}-\mathrm{c}_i \right\|_2
\end{equation}
where $i \in \{1, \cdots, K\}$ and a corresponding weight vector $\omega$ for each feature is associated with one of codewords based on the assigned number.
\begin{equation}
\omega_{\mathrm{x}_{HA}} = \mathrm{c}_i, \quad \text{where  } i = A_{HA_i}.
\end{equation}
HA encoding enables reducing memory requirements by maintaining only codewords and the assigned codeword numbers instead of keeping all features. Moreover, the codeword can be ignored and initialized with random values when learning a sequence of assigned numbers. Thus, a video can be represented by a sequence of assigned numbers, leading to memory saving.

\textbf{- Soft Assignment}: The SA method considers $k$-nearest codewords to the feature. Fig.~\ref{fig:5nn_vq} illustrates an example of 5 nearest neighbor (NN) codewords ($5$-SA). Five red nearest codewords are correlated with the feature vector $\mathrm{x}$ and a weighted centroid vector colored in green is then computed for assignment. The weight vector $\omega$ is computed as follows.
\begin{equation}
\omega_{\mathrm{x}_{SA}} = \sum_{j=1}^{K}\delta(\mathrm{x},\mathrm{c}_j) \cdot  \mathrm{c}_j \cdot d_{\omega_j}
\label{equ:SA_weight}
\end{equation}
where $d_{\omega_j}$ is the normalized inverse distance weight:
\begin{equation}
d_{\omega_j} =
\frac{\delta(\mathrm{x},\mathrm{c}_j)\exp(-\beta\left\| \mathrm{x}-\mathrm{c}_j\right\|_2^2)}{\sum_{j=1}^{K}\delta(\mathrm{x},\mathrm{c}_j)\exp(-\beta\left\| \mathrm{x}-\mathrm{c}_j\right\|_2^2)}
\end{equation}
where $\delta(\mathrm{x},\mathrm{c}_j)$ is the indicator function for the $k$-NN codewords of $\mathrm{x}$:
\begin{align}
	\delta(\mathrm{x},\mathrm{c}_j) = 
	\begin{cases}
		1, & \text{if } \mathrm{c}_i \in k\text{-NN}(\mathrm{x}),  \\
		0, & \text{otherwise.}
	\end{cases}
\end{align}
Thus, the computed weight vector $\omega$ gives the weighted centroid of $k$-NN codewords based on inverse distance between the feature and $k$ nearest codewords.
Each weight vector $\omega$ is unique, and therefore an assigned number for each weight vector $\omega$ is also unique. 
Hence, the total number of assigned numbers is the same as the total number of extracted features in a dataset.

\begin{align}
	A_{SA_i} = i, \quad \text{where } i \in \{1, \cdots, \mathrm{N}\}.
\end{align}
When learning an \textit{ActionWord} encoded with SA, random vector initialization of the weight vectors cannot be feasible as the assigned numbers are nothing but sequential numbers for each feature.
Note that HA can be regarded as a special case of $k$-SA, where $k$ is 1.

\textbf{- Direct Assignment}: Instead of computing the codebook, Direct Assignment (DA) encoding considers each video-frame feature as a weighted codeword and assign a unique number to it. 
\begin{align}
	\omega_{\mathrm{x}_{DA}} &= \mathrm{x} \\
	%\label{equ:DA_weight}
	%\end{equation}
	%\begin{align}
	A_{DA_i} &= i, \quad \text{where } i \in \{1, \cdots, \mathrm{N}\}.
	\label{equ:DA_assign}
	%\end{align}
\end{align}
Each frame feature vector is thus directly considered as an \textit{ActionWord}. This method does not require codebook generation leading to reduced computation time, but the memory requirement increases.

%-------------------------------------------------------------------------
\subsection{Sequence Learning with Temporal ConvNet} \label{apporach:learn_video}

With the proposed \textit{ActionWord} coding, action prediction and action recognition can be regarded as classification problems for a partial sentence or a sentence. By leveraging the success of sentence classification using ConvNets~\cite{text_ref1, text_ref3, text_ref4, text_ref5} in NLP, we apply similar ConvNet architectures to train and classify \textit{ActionWord} sequences.
We consider two ConvNet models: $\mathrm{i})$ T-CNN, $\mathrm{ii})$ Covolutional LSTM (C-LSTM).

\begin{figure}[!tbp]
	\captionsetup[subfigure]{justification=centering}
	\centering
	\begin{subfigure}[b]{.22\textwidth}
		\centering
		\includegraphics[width=0.8\linewidth]{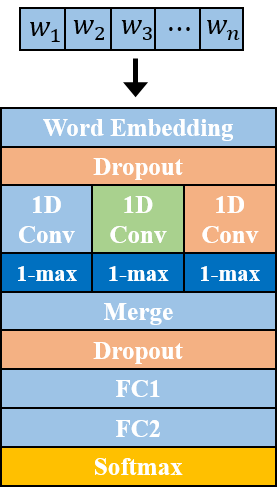}
		\caption{T-CNN Model}
		\label{fig:cnn_model}
	\end{subfigure}%
	\begin{subfigure}[b]{.22\textwidth}
		\centering
		\includegraphics[width=0.8\linewidth]{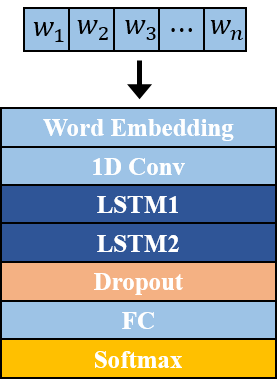}
		\caption{Covolutional LSTM Model}
		\label{fig:clstm_model}
	\end{subfigure}
	\caption{ConvNet Architectures}
	\label{fig:learning_models}
	\vspace*{-0.5cm}
\end{figure}

\noindent \textbf{Word Embedding}: The sequence of \textit{ActionWord}s is the input to the ConvNets shown in Fig.~\ref{fig:learning_models}.
Since the length of the sequence for each video is different, a word embedding layer is utilized to make the sequences of the same length. The length of each sequence $l_i$ is truncated if $l_i > l_{max}$ whereas $l_i$ is padded with a special codeword that corresponds to $v=[0, \cdots,0]$ if $l_i < l_{max}$, where $v\in \mathbb{R}^D$ and $l_{max}$ is a user-determined sequence length.
The word embedding layer combines the corresponding weight vector $\omega$ based on the assigned word number, and generates an $D \times l_{max} $ matrix for each sequence.
The weight vector can be initialized with a random number between -0.05 and 0.05 for the HA random initialization encoding method.
%The random weight vector can be learned in a new vector representation while training. 

\noindent \textbf{T-CNN Model}: Fig.~\ref{fig:cnn_model} shows the overall structure of the T-CNN Model.
T-CNN consists of $L$ one-dimensional convolution layers denoted by $C^{l} \in \mathbb{R}^{F_l\times T}$ in parallel where $F_l$ is the number of convolution filters in the $l$-th layer and $T$ is same as $l_{max}$. Each layer consists of temporal convolution, a non-linear activation, and global max (1-max) pooling across time. The collection of filters in each layer is defined as $W= \{W^{(i)}\}^{F_l}_{i=1}$ where $W^{(i)} \in \mathbb{R}^{d \times F_{l}}$ and a window of $d$ duration. The corresponding bias vector is $b \in \mathbb{R}^{F_l}$. Given the input sequence of weight vectors, $\Omega \in \mathbb{R}^{D \times T}$, the activation $C^{l}$ is computed such that
\begin{equation}
C^{l} = ReLU(\mathbb{W} \ast \Omega + b)
\label{equ:T-CNN_conv}
\end{equation}
where $\ast$ is the convolution operator. The convoluted signals can be viewed as $N$-gram  in a sentence, where $N$ can be determined by the size of filters in the convolution layer. After the ReLU activation, the global max pooling is applied to get the largest signal from the activation. Each layer produces a $\nu$ vector where $\nu \in \mathbb{R}^{F_l}$ by concatenating the global max signals. All vectors from $L$ layers are then concatenated generating a $v$ vector where $v=\sum_{i=1}^{L}\nu_i$. The output size of the second FC layer is the number of class in a dataset and Softmax activation is applied in the end.

\noindent \textbf{Covolutional LSTM Model}: C-LSTM Model consists of a convolution layer and a long short-term memory recurrent neural network (LSTM)~\cite{related_ref1} designed for time-series data to learn long-term information.
%and hence we add the LSTM layer on top of the CNN layer to learn higher-level features.
Fig.~\ref{fig:clstm_model} shows the overall architecture of the C-LSTM.
%As shown in Fig.~\ref{fig:clstm_model}, the first layer of the model, the one dimensional convolution layer, is same as the one with the CNN model. 
The multiple parallel convolution layer is not applied because the concatenation of the resulting vectors can break the original sequence for the input of the LSTM layer. 
The global max pooling layer is also omitted for the same reason. 
We retain the original order of the sequence and extract more descriptive representations by convolution computation for the sequence.
The extracted local temporal information is fed into the LSTM layer and the LSTM layer outputs a video level representation that captures high level temporal information.

%------------------------------------------------------------------------
\section{Experiments} \label{experiments}

\begin{table}[t]                                               
	\centering                                                  
	\begin{tabu} spread \linewidth {|X[0.8cm]||X[cm]|X[cm]|} %to 1.0\linewidth [m]{|X[m,c]||X[m,c]|X[m,c]|}
		\hline                  
		%		\rowfont{\bfseries}                                    
		& UCF101 & HMDB51  \\ 
		\hline  
		\hline                                                    
		$C$  & 101 & 51 \\        
		\hline                                                      
		$l_{train}$ & 35.8 (4 / 354) & 17.7 (2 / 211) \\         
		\hline  
		$l_{test}$  & 35.3 (4 / 177) & 17.1 (3 / 128) \\         
		\hline 
		$N$ & 9537 (3783)  & 3570 (1530) \\         		
		\hline
		
	\end{tabu}                                               
	\caption{Summary statistics of extracted features for each dataset. $C$: number of classes, $l_{train}$: average sequence length of training data (min / max), $l_{test}$: average sequence length of testing data (min / max), $N$: number of training(testing) sequences(or videos) for each dataset}
	\label{table:stats}                                  
	\vspace*{-0.5cm}
\end{table}  

\subsection{Dataset and Statistics}
We test our method on two action video datasets, HMDB51~\cite{ar_ref3} and UCF101~\cite{ar_ref4}. The HMDB51 dataset consists of 51 action classes with 6,766 videos and more than 100 videos in each class. All videos are acquired from movies or Youtube, and contain various human activities, including interactions with other humans or objects. Each action class has 70 videos for training and 30 videos for testing. 
The UCF101 dataset consists of 101 action categories with 13,320 videos and at least 100 videos are involved in each class. All videos are gathered from Youtube.

Both datasets provide three training and testing splits. We used the first split of each dataset for validating our proposed models. The same parameters and models from split 1 are utilized for other two splits.
%Since we represent the videos with sequences of \textit{ActionWord}, 
Table~\ref{table:stats} shows the statistics of sequence lengths on each dataset for our experiments. We extracted temporal features every 5 frames ($\mathrm{T}=5$) with 10 stacked input frames ($\mathrm{L}=10$) and spatial features every 5 frames.

%-------------------------------------------------------------------------
\subsection{Implementation Details}

\noindent \textbf{Training Two-ConvNets}: We use the VGG-16 model~\cite{ar_ref2} for two-stream ConvNets training.
Both the temporal and the spatial network are initialized with the pre-trained weights trained with ImageNet~\cite{dataset1}. The networks are then fine-tuned with each dataset.

For the training of the spatial network, we use dropout ratios of 0.8 for two FC layers. The input images are resized to make the smaller side as 256. We augment the input images by randomly cropping 224$\times$224 sub-images from the four corners and the center of the original images and randomly flipping in horizontal direction. The learning rate is set to $10^{-3}$ initially and decreased by a factor of 10 when the validation error saturates.

For the training of the temporal network, we use dropout ratios of 0.9 for UCF101 and 0.9 and 0.8 for HMDB51. We pre-compute the optical flows using the TVL1 method~\cite{dataset2} before training to improve the training speed. The optical flow input is stacked with $\mathrm{L}=10$ frames making a 224$\times$224$\times$20 sub-volume. Same data augmentation techniques are employed for the sub-volume and the learning rate is initialized with $5\times10^{-4}$ and decreased in the same manner of the spatial network training. A mini-batch of 128 samples are employed at each iteration, but batch normalization method~\cite{related_ref5} is not used for all trainings.

\noindent \textbf{Word Vector Representation}: The dimension of temporal $\mathrm{x}_t$ and spatial $\mathrm{x}_s$ feature vectors is 4096. 
Since the two extracted feature vectors are complementary, we concatenate them with a data ratio $r$, resulting in a combined feature vector $\mathrm{x}$.
\begin{align}
	\label{equ:data_ratio}
	\mathrm{x} = PCA(\mathrm{x}_{t(1:rD)}) \oplus PCA(\mathrm{x}_{s(1:(1-r)D)})
\end{align}
where $D$ is the dimension of $\mathrm{x}$, $0\leq r \leq 1$,  $\oplus$ is a concatenation operation, and $PCA(\mathrm{x}_{1:n})$ is to apply PCA to $\mathrm{x}$ and take the first $n$ elements of the projected vector. The reduced dimension of $\mathrm{x}$ is $D'\in\{32,64,128,256,512,1024\}$.
We use the output vector of the penultimate FC (FC7) layer, since the performance with the FC7 vectors is consistently  2$\sim$3\% better than the one with the first FC (FC6) layer.
In addition, we take the output vector of FC7 with input images or optical flow images that are cropped in the center area making size of 224$\times$224.
For the SA and HA feature encoding method, we consider $K=\{5000, 10000, 20000\}$ as the size of codebook.

\noindent \textbf{Training T-CNN Model for Sequence Learning}: We use three ($L$=3) parallel 1D convolution layers whose filter sizes are 3,4,5 respectively and number of filters are 200.
The first dropout rate and the second one are 0.2 and 0.8, respectively. Since the model is simple, we use a somewhat strong dropout rate to prevent from overfitting.
The T-CNN model is trained with a mini-batch size of 64 and the training is terminated after 100 and 300 epochs for UCF101 and HMDB51, respectively.

\noindent \textbf{Training C-LSTM Model for Sequence Learning}: The filter size of the 1D convolution layer is 5 and its filter count is 200.
The number of hidden units of the first and second LSTM layers is 100 and the dropout rate is set to 0.6. Training is terminated after 100 and 200 epochs for UCF101 and HMDB51, respectively. For both models, we use categorical cross entropy loss with Stochastic Grandient Descent and RMSProp ~\cite{text_ref7} step updates, whose learning rate is initialized with $10^{-4}$. 

\noindent \textbf{Tesing}: Given the trained models (T-CNN, C-LSTM), we evaluate the accuracy with the full sequences for the action recognition task, as well as partial sequences for action prediction. Each video sequence is divided into 10 segments creating the following sequences for action predection~\cite{ap_ref2, ap_ref4, ap_ref5, ap_ref6}: 0$\sim$10\%, 0$\sim$20\%, $\cdots$, 0$\sim$100\%.

\noindent \textbf{Running Time}: The running time of our method is compared with MTSSVM~\cite{ap_ref4}, MSSC~\cite{ap_ref1}, and Two-stream Fusion~\cite{ar_ref5} methods and the results are listed in Table~\ref{table:running_time}. We executed authors' code on a 4.6GHz CPU with 32GB RAM and one TITAN-X GPU. 
With a sequence of 512-dimension weight vectors, the training time is 51min(T-CNN) and 101min(C-LSTM) on UCF101, and 10min(T-CNN) and 67min(C-LSTM) on HMDB51. Note that the testing time takes a few seconds for each dataset. The T-CNN method is 170$\times$, 507$\times$, 425$\times$ faster than MTSSVM, MSSC, Fusion methods, respectively on UCF101. For the HMDB51 dataset, the T-CNN method is 377$\times$, 1150$\times$, 945$\times$ faster than MTSSVM, MSSC, Fusion methods, respectively. The C-LSTM method also spends much less time than compared methods. Note that training time of two-stream ConvNet and feature extraction is not included.

\begin{table}[!thb]                                  
	\centering                                     
	\begin{tabular}{|>{\centering\arraybackslash}m{3.1cm}||>{\centering\arraybackslash}m{2.0cm}|>{\centering\arraybackslash}m{2.2cm}|} 
		\hline      
		Methods & UCF101 (hrs) & HMDB51 (hrs) \\
		\hline \hline 
		MTSSVM~\cite{ap_ref4} & 145 & 83 \\                    
		\hline                                         
		MSSC~\cite{ap_ref1} & 431 & 253 \\
		\hline                                         
		Fusion~\cite{ar_ref5} (15 epoch) & 362 & 208 \\                  
		\hline                                         
		Ours (T-CNN) & 0.85 & 0.22 \\                  
		\hline                                         
		Ours (C-LSTM) & 1.68 & 1.12 \\                
		\hline
	\end{tabular}                                  
	\caption{Training and testing time of comparison methods in hours on UCF101 and HMDB51.}
	\label{table:running_time}
\end{table}

\subsection{Baseline of Two-stream ConvNets}
Table~\ref{table:baseline} shows baseline accuracies for the spatial, temporal, two-stream  networks on UCF101 and HMDB51. The value is averaged over three splits and two-stream results are obtain by averaging the prediction probabilities of the spatial and temporal ConvNets. The proposed methods leverage thes baseline two-strema ConvNet and show improvement by taking the temporal information into account.

\begin{table}[!thb]                                  
	\centering                                     
	\begin{tabular}{|>{\centering\arraybackslash}m{2.0cm}||>{\centering\arraybackslash}m{2.0cm}|>{\centering\arraybackslash}m{2.0cm}|} 
		\hline      
		& UCF101 & HMDB51 \\
		\hline \hline 
		Spatial & 81.8 & 44.8 \\                    
		\hline                                         
		Temporal & 84.9 & 55.0 \\
		\hline                                         
		Two-stream & 90.1 & 61.4 \\                  
		\hline
	\end{tabular}                                  
	\caption{Baseline mean performance of spatial, temporal, and two-stream ConvNet on UCF101 and HMDB51. (VGG-16 CNN model is employed.)}
	\label{table:baseline}
\end{table}

\subsection{Parameter Analysis}

\begin{figure}[!tb]
	\begin{center}
		\includegraphics[width=1.0\linewidth]{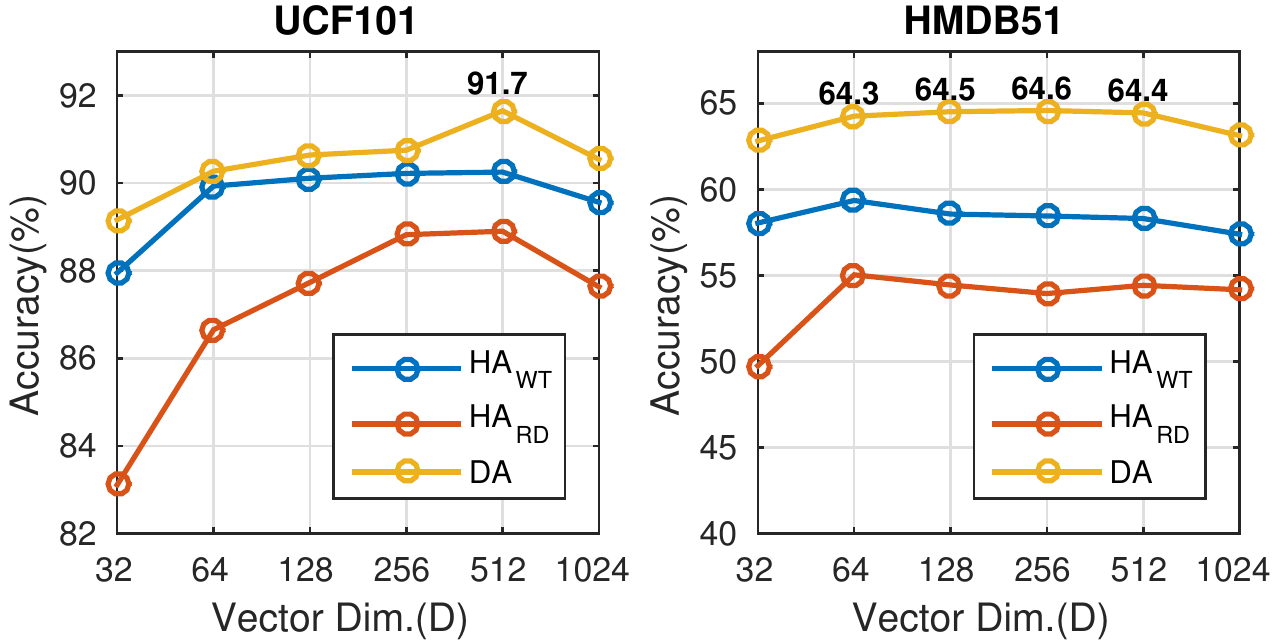} % fig_vecdim_weight_anal.eps
	\end{center}
	\caption{Accuracy based on different initialization and dimension of the weight vector $\omega$. $HA_{RD}$ and $HA_{WT}$ denote random initialization and assigned codebook initialization, respectively.}
	\label{fig:vecdim_weight_vs_acc}
\end{figure}

\noindent \textbf{Effects of Dimension and Initialization of Weight Vector}:
%We first investigate the performance based on different weight vector initialization 
We first investigate how the weight vector initialization and feature vector size affect the performance. We experiment by setting parameters: with equal data ratios ($r=0.5$) for temporal and spatial features, with full testing sequences, and with $K=20$k.
Fig.~\ref{fig:vecdim_weight_vs_acc} shows the results with the T-CNN model. The vectors initialized with weight vectors outperforms randomly initialized weight vectors on both datasets and the performance margin is smaller, as the vector size increases. The randomly initialized vector takes about twice more epochs to be fully trained but data storage can be saved substantially.

In addition, the performance on UCF101 increases as the feature vector dimension increases until 512 with both HA and DA. We speculate this trend occurs because more data is generally helpful but data of size larger than 512 can contain less important data from PCA, so the performance is degraded thereafter. Similar trend happens on the HMDB51 dataset, but no significant performance change is observed between feature vectors of 64 and 512. 
This means that our method is robust to the choice of the vector dimension results except the 32-dim vector which loses too much information.

\begin{figure}[!tb]
	\begin{center}
		\includegraphics[width=1.0\linewidth]{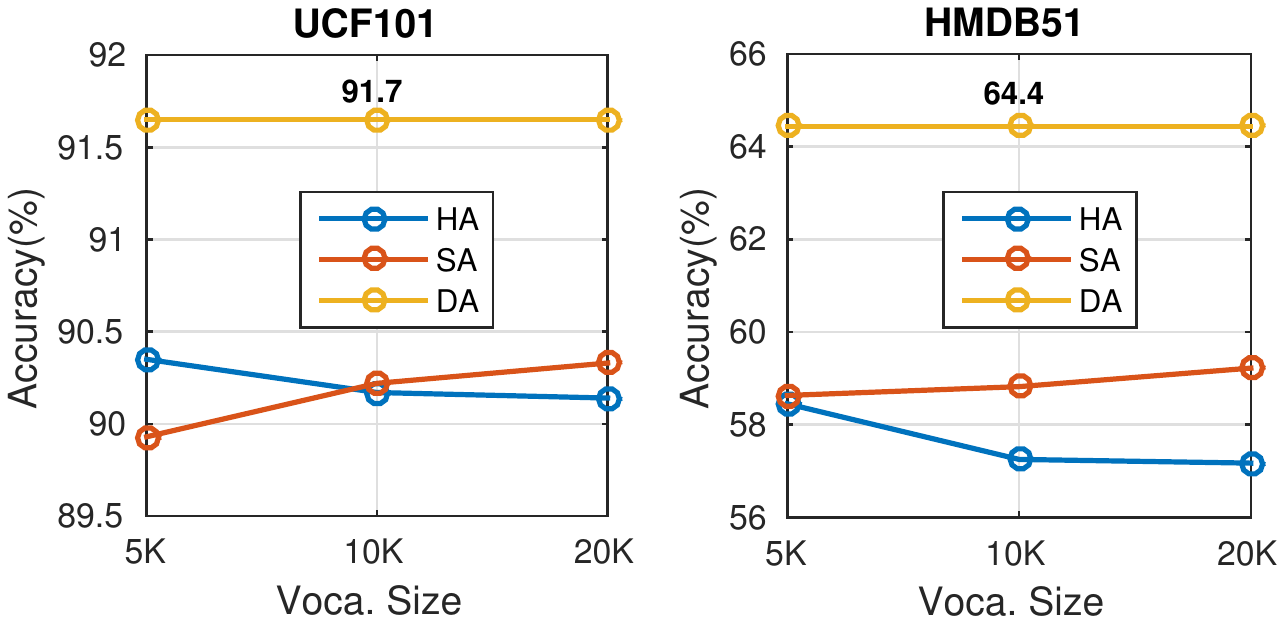} % fig_voca_assign_anal.eps
	\end{center}
	\caption{Accuracy based on different size of codebook and different encoding methods.}
	\label{fig:vocasize_encoding_vs_acc}
\end{figure}

\begin{figure}[!tb]
	\centering
	\begin{subfigure}{0.49\linewidth}
		\centering
		\includegraphics[width=1\linewidth]{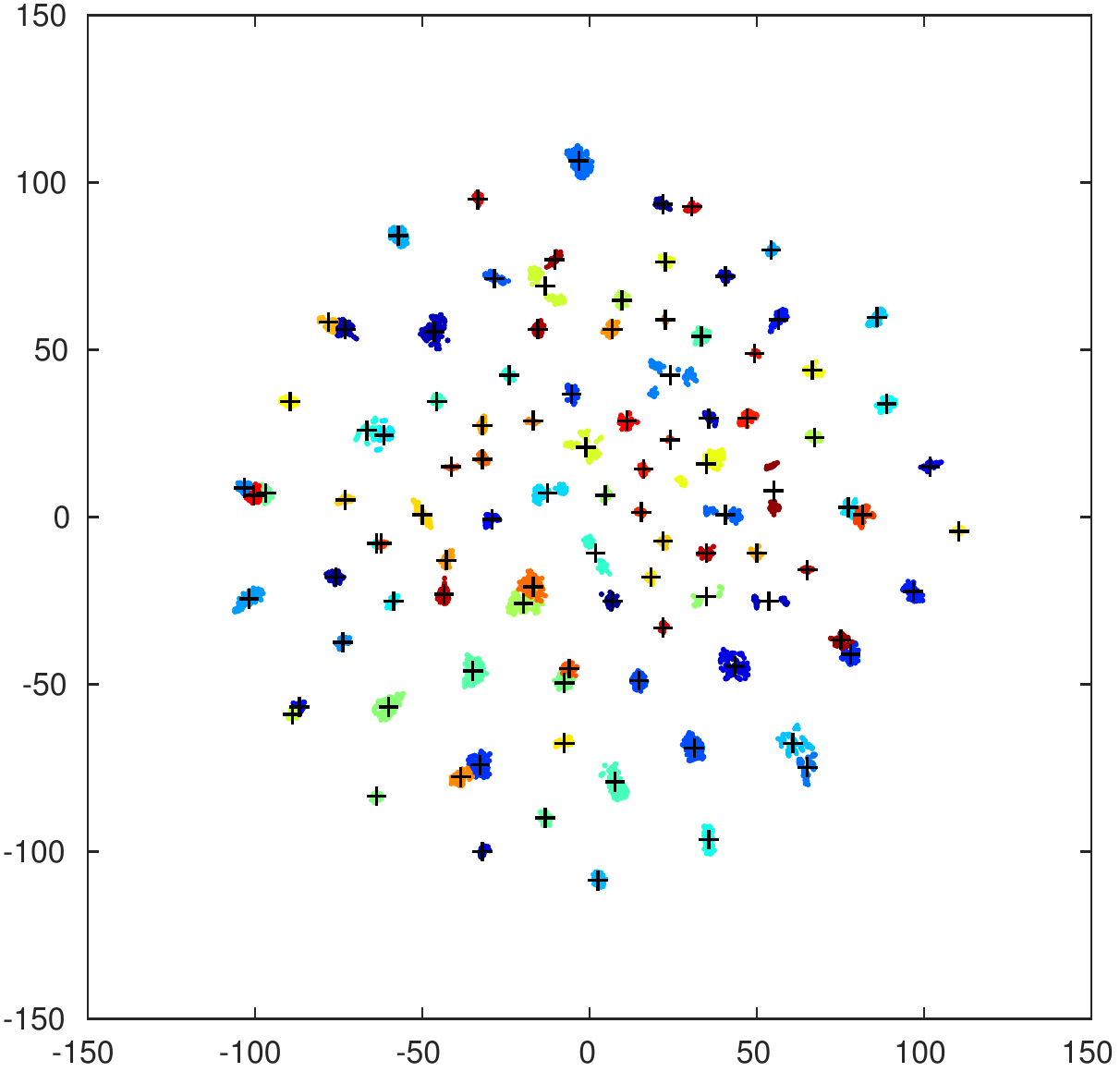} %fig_tsne_5k_128_to2_GMM_UCF101.eps}
		\caption{5k Codebook}
		\label{fig:5k_codebook}
	\end{subfigure}	
	\begin{subfigure}{0.49\linewidth}
		\centering
		\includegraphics[width=1\linewidth]{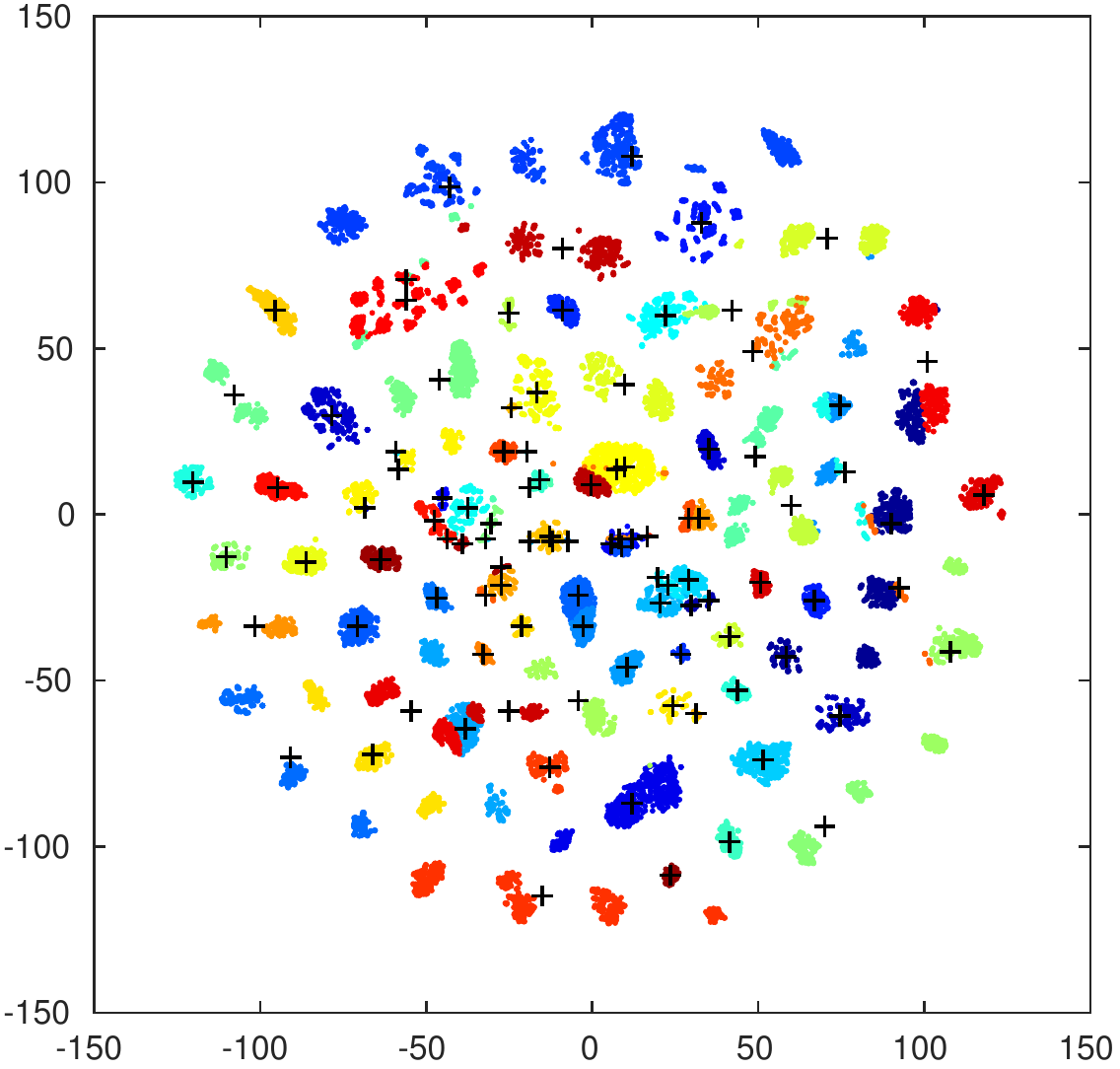} %fig_tsne_20k_128_to2_GMM_UCF101.eps}
		\caption{20k Codebook}
		\label{fig:20k_codebook}
	\end{subfigure}
	\caption{Visualization of 5k and 20k codebooks ($D=2$) of UCF101 c. Each codebook is clustered with $k$-means ($k=101$).}
	\label{fig:tsne_codebook_mapping}
\end{figure}

\noindent \textbf{Effects of Codebook Size and Encoding Methods}.
In this experiment, we observe the performance given different codebook sizes and encoding methods. The dimension of the feature vector is fixed to 512, since in the previous experiment the size 512 is found as the most optimal length. The data ratio $r$ is set to 0.5.
Fig.~\ref{fig:vocasize_encoding_vs_acc} shows the results with the T-CNN model. The performance of HA decreases as the codebook size increases, while the SA performance increases with larger codebook. In order to investigate these trends, we reduce 128-dimensional 5k and 20k codebooks on UCF101 to 2-dimensional vectors respectively and cluster them with $k$-means, where $k=101$. We employ the t-SNE dimensionality reduction technique ~\cite{related_ref4}, which is well suited for displaying high-dimensional data. As shown in Fig.~\ref{fig:tsne_codebook_mapping}, the 5k codebook has larger margin between clusters than the 20k codebook. Therefore, with HA, it is less likely to mislabel with the 5k codebook than the 20k codebook. On the other hand, with SA, the 5 NN codebooks can group more tightly with the 20k codebook, so the centroid of 5NN is likely to be closer to the original feature vector than the centroid in the 5k codebook.
In any cases, since the performance gain of different codebook sizes is small, we can argue that our method is robust to the choice of the codebook size.
Another distinctive observation is that DA outperforms other encoding methods with relatively large margin.

\subsection{Optimal Data Ratio} \label{sec:opt_data_ratio}

\begin{figure}[!tb]
	\centering
	\begin{subfigure}[b]{.49\textwidth}
		\centering
		\includegraphics[width=\linewidth]{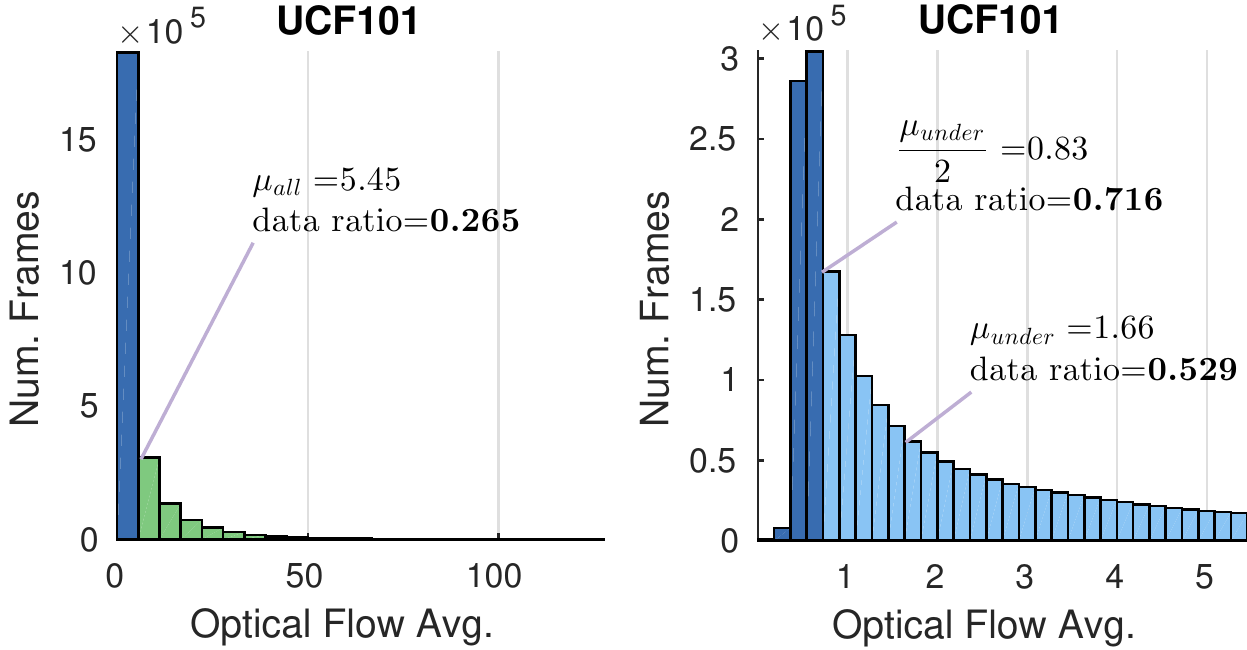} % fig_flow_anal_ucf101.eps
		%\caption{}
		\label{fig:flow_UCF101}
	\end{subfigure}		
	
	\begin{subfigure}[b]{.47\textwidth}
		\centering
		\includegraphics[width=\linewidth]{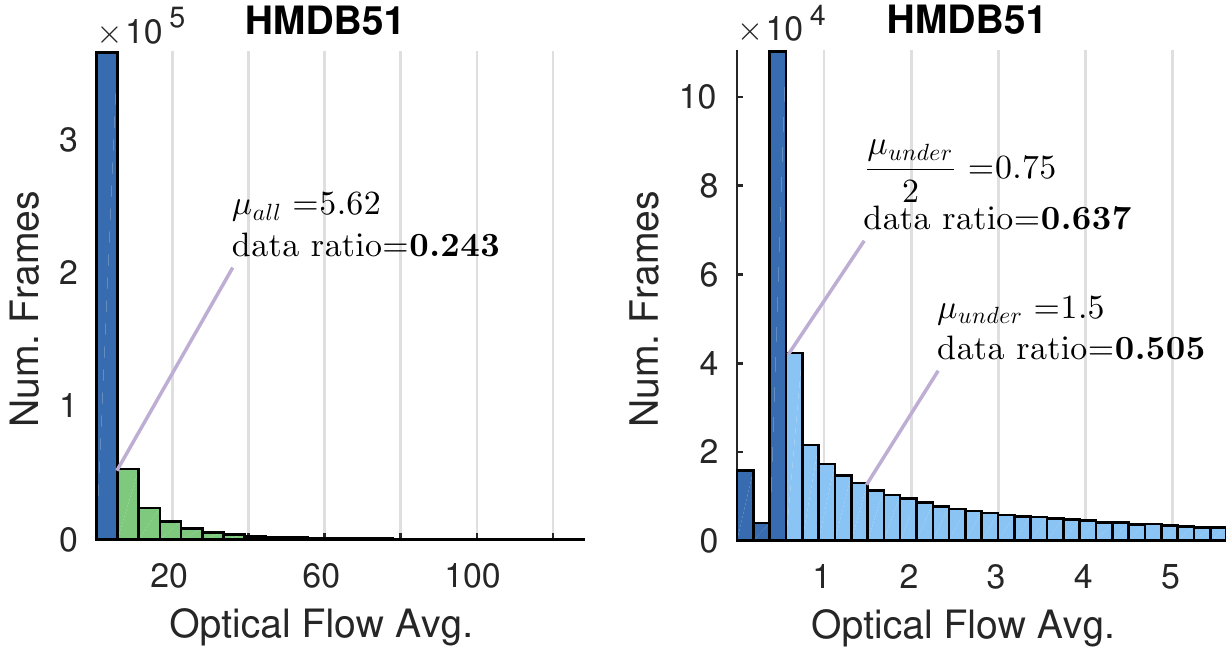} % fig_flow_anal_hmdb51.eps
		%\caption{}
		\label{fig:flow_HMDB51}
	\end{subfigure}
	\caption{Histogram of average optical flow on UCF101 and HMDB51.}
	\label{fig:optical_flow_stats}
\end{figure}

\begin{table}                                            
	\centering                                               
	\begin{tabular}{|>{\centering\arraybackslash}m{2cm}||>{\centering\arraybackslash}m{1.5cm}|>{\centering\arraybackslash}m{1.5cm}|>{\centering\arraybackslash}m{1.5cm}|}    
		\hline
		HMDB51 & $r=0.5$ & $r=0.625$ & $r=0.75$ \\                                 
		%\hline                                          
		%64 (0-1) & 64.2 & 65.0 & 64.3 \\                         
		\hline                                                   
		64 & 65.2 & 65.2 & 66.0 \\ %(0.1-1)
		%\hline                                                   
		%128 (0-1) & 64.5 & 65.3 & 64.6 \\                        
		\hline                                                   
		128 & 65.0 & 65.6 & 65.0 \\ %(0.1-1)
		%\hline                                                   
		%256 (0-1) & 64.6 & 64.7 & 63.9 \\
		\hline                                                   
		256 & 64.6 & 65.7 & 64.6 \\ %(0.1-1)
		%\hline                                                   
		%512 (0-1) & 64.4 & 65.5 & 64.7 \\                        
		\hline                                                   
		512  & 64.8 & \textbf{66.4} & 65.1 \\ %(0.1-1)
		\hline  \hline                       
		UCF101 & $r=0.5$ & $r=0.625$ & $r=0.75$ \\                           
		\hline
		512 & 91.5 & 91.8 & \textbf{92.7} \\ %(0-1)
		\hline                                                   
	\end{tabular}                                            
	\caption{Performance based on different data ratios and feature dimensions on HMDB51 and UCF101 split 1.}
	\label{table:dataRatio_flowRatio_dataset}
	\vspace*{-0.3cm}	                    
\end{table}      

The temporal and spatial feature vectors are concatenated based on the data ratio $r$ in eq. (\ref{equ:data_ratio}). As shown in Table~\ref{table:baseline}, the temporal network outperforms the spatial network on both datasets. In this analysis, we empirically find an optimal ratio that assigns higher weight to the temporal feature vector.

First, we compute the frame-wise average of optical flow magnitudes along the two axis as follows:
\begin{align*}
	f_i = \frac{1}{2} \left(  \nicefrac{\sum_{k=1}^{P}{abs(f_{u_{i,k}}-128)}}{P} + \nicefrac{\sum_{k=1}^{P}{abs(f_{v_{i,k}}-128)}}{P} \right) 
\end{align*}
where $f_i$ is the average optical flow for the i-th frame in the video, $P$ is the total number of pixels in the i-th frame, and $f_u$, $f_v$ are the horizontal and the vertical optical flow values, respectively. Of course, the intuition is that frames with higher motion information can be identified using $f_i$.

\begin{table}[!tb]                             
	\centering                                
	%\begin{tabular}{|c|c|c|c|}                
	\begin{tabu} to 1.0\linewidth {|X[1.8c]X[0.5c]|X[1.8c]X[0.5c]|}  
		\hline 
		\multicolumn{2}{|c|}{HMDB51} & \multicolumn{2}{c|}{UCF101} \\                  
		\hline \hline                              
		iDT+FV~\cite{ar_ref9} & 57.2 & iDT+FV~\cite{ar_ref10} & 85.9 \\             
		\hline                                    
		iDT+HSV~\cite{bovw_ref8} & 61.1 & iDT+HSV~\cite{bovw_ref8} & 87.9 \\            
		\hline
		VideoDarwin~\cite{ar_ref6} & 63.7 & LRCN~\cite{ar_ref8} & 82.9 \\        
		\hline                                    
		Two stream~\cite{ar_ref1} & 59.4 & Two stream~\cite{ar_ref1} & 88.0 \\         
		\hline                                    
		TDD+FV~\cite{ar_ref7} & 63.2 & TDD+FV~\cite{ar_ref7} & 90.3 \\             
		\hline                                    
		KVMF~\cite{ar_ref11} & 63.3 & KVMF~\cite{ar_ref11} & \textbf{93.1} \\               
		\hline                                    
		Fusion~\cite{ar_ref5} & {65.4} & Fusion~\cite{ar_ref5} & \textbf{92.5} \\             
		\hline                                    
		Transformation~\cite{ad_ref4} & 62.0 & Transformation~\cite{ad_ref4} & 92.4 \\     
		\hline  \hline                                   
		Ours(C-LSTM) & 62.4 & Ours(C-LSTM) & 90.9 \\      
		\hline                                    
		Ours(T-CNN) & \textbf{66.3} & Ours(CNN) & \textbf{92.5} \\         
		\hline                                    
	\end{tabu}                             
	\caption{Action recognition performance comparison with State-of-the-art. (mean over three splits)}
	\label{table:state_art}   
	\vspace*{-0.5cm}	                    
\end{table} 

\begin{table*}[!thb]                                                                       
	\centering 
	\begin{tabu} to 1.04\textwidth {|X[1.55c]||X[1c]X[1.1c]X[1.1c]X[1.1c]X[1.1c]X[1.1c]X[1.1c]X[1.1c]X[1.1c]X[1.28c]|}  
		\hline
		UCF101 & 0-10\% & 0-20\% & 0-30\% & 0-40\% & 0-50\% & 0-60\% & 0-70\% & 0-80\% & 0-90\% & 0-100\% \\                
		\hline 
		MOS~\cite{ap_ref6} & \textendash & 35.0 & \textendash & 37.1 & \textendash & 39.4 & \textendash & 40.3 & \textendash & 40.9 \\                             
		\hline
		SMMED\cite{ap_ref6} & \textendash & 40.6 & \textendash & 40.6 & \textendash & 40.6 & \textendash & 40.6 & \textendash & 40.6 \\                           
		\hline
		Fusion~\cite{ar_ref5} & \textbf{82.8} & 85.5 & 87.5 & 88.8 & 89.2 & 90.4 & 90.7 & 91.0 & 91.5 & 92.5 \\
		\hline
		Ours & 82.2 & \textbf{86.7} & \textbf{88.5} & \textbf{89.5} & \textbf{90.1} & \textbf{91.0} & \textbf{91.5} & \textbf{91.9} & \textbf{92.4} & \textbf{92.5} \\             
		\hline \hline
		HMDB51 & 0-10\% & 0-20\% & 0-30\% & 0-40\% & 0-50\% & 0-60\% & 0-70\% & 0-80\% & 0-90\% & 0-100\% \\
		\hline 
		Fusion~\cite{ar_ref5} & \textbf{44.8} & 51.5 & 54.5 & 58.0 & 61.0 & 62.9 & 64.9 & 65.2 & 65.4 & 65.4 \\ % \textendash
		\hline
		Ours & 38.8 & \textbf{51.6} & \textbf{57.6} & \textbf{60.5} & \textbf{62.9} & \textbf{64.6} & \textbf{65.6} & \textbf{66.2} & \textbf{66.3} & \textbf{66.3} \\             
		\hline 
	\end{tabu}
	\caption{Action Prediction performance on UCF101 and HMDB51.}
	\label{table:action_prediction}
	\vspace*{-0.3cm}	                    
\end{table*}

We explain the choice of $r$ using the histograms of $f_i$ shown in Fig.~\ref{fig:optical_flow_stats}. The left column shows that the frames in the green colored bins contain more motion cues than the frames in the blue colored bins. Also, majority of the frames fall below the mean of the $f_i$ across all frames, i.e. $\mu_{all}$. These are frames that contain less motion information, and hence provide more spatial appearance information. A first order estimate of $r$ could then be given by the ratio of frames above $\mu_{all}$ over total number of frames. However, since motion is a stronger cue,it is reasonable to assume that better estimates of $r$ would be given by the first quartile or the half of the first quartile. Therefore, consider the graphs on the right column of Fig.~\ref{fig:optical_flow_stats}, which show the histograms of $f_i$ only for frames whose average optical flow is smaller than $\mu_{all}$. We compute the mean of these lower histograms, denoted as $\mu_{under}$, which determine the first quartile of the original histogram. Better estimates of the ratio $r$ are then given by the ratio of frames above $\mu_{under}$ or $\mu_{under}/2$ over the total number of frames. 

In our experiments, we found that the ratio $r$ given by $\mu_{under}$ is 0.529 on UCF101 and 0.505 on HMDB51 meaning that $\mu_{under}$ is close to median of the average optical flows. The estimate based on $\nicefrac{\mu_{under}}{2}$, resulted in $\sim$0.75 for UCF101 and $\sim$0.625 for HMDB51. One observation is that the UCF101 dataset involves many sports and exercise videos~\cite{ad_ref3} that generally contain larger motions, while the HMDB51 dataset consists of simple action videos~\cite{ad_ref3} that have moderate motion. The ratios computed with $\nicefrac{\mu_{under}}{2}$ support this observation. 
Results using DA and T-CNN for these ratios are shown in Table~\ref{table:dataRatio_flowRatio_dataset}.
The best performance is achieved with estimates based on $\nicefrac{\mu_{under}}{2}$, confirming that the estimated ratios are reliable.

\subsection{Action Recognition Performance} \label{sec:state}

Table~\ref{table:state_art} shows action recognition results of recent state-of-the-art methods. Our best result outperforms other methods by 0.9\% on HMDB51 and is compatible on UCF101. We conjecture that \cite{ar_ref11} outperforms ours because they utilize GoogLeNet~\cite{related_ref6} with batch normalization~\cite{related_ref5}, which is a deeper network than VGG-16~\cite{ar_ref2}. Our result is on par with Fusion~\cite{ar_ref5} on UCF101 but its computational efficiency is much better due to the fast-trainable network as shown in Table~\ref{table:running_time}.
The C-LSTM model, however, does not learn much comparing with the baseline accuracy. We speculate this is because the temporal 1D convolution without pooling does not represent a video effectively. Applying 1D convolution followed by max pooling over several small segments may boost the performance for the C-LSTM model.

\subsection{Action Prediction Performance} \label{sec:ap}

The goal in action prediction is the same as in action recognition, except that the input test video is not a full video. Our method can take a variable size input so the partial input can be readily handled.
In order to compare with a method using T-CNN, we evaluate Fusion~\cite{ar_ref5} with the partial test video frames. We follow their testing procedure by taking 5 uniformly spaced frames from the given range. The horizontally flipped input frame is augmented and the entire frame is used.

Table~\ref{table:action_prediction} show the action prediction results with comparing methods. Our results consistently outperform the Fusion method as well as the previous best results: MOS and SMMED~\cite{ar_ref6}. 
%Note that the best trained T-CNN model for HMDB51 is obtained by discarding the beginning 10\% portion. Thus, a CNN model trained with the full sequence is also employed for the sequence starting from the beginning.
%as we found the beset sequence length for HMDB51 starts from 10\% of the video the beginning 10\% of the sequence input is dropped out except the 0-10\% range. We adopt the same way for the evaluation of the Fusion method.
We observe an interesting trend, in the sense that our result is only outperformed by Fusion in the first 10\% range. We conjecture two reasons about the result: the length of the sequence is too short to be fully trained, and noisy words are inserted to the sequence especially on HMDB51. 
On the other hand, our method rapidly reaches to full accuracy with partial data. The prediction results with half-video data reach 95\% and 97\% of full accuracy for the HMDB51 and UCF101, respectively. Also, the performance with 90\% of frames is almost identical to full accuracy. These observations show that our method is well suitable to detect actions with partial data.

%------------------------------------------------------------------------
\section{Conclusion} \label{conclusion}

We proposed an effective and efficient sequence learning method that captures global temporal sequencing information of a video. This is achieved by means of a new video representation as a sequence of visual words (a sentence). By training a ConvNet to learn the sequences corresponding to different actions, we are able to accurately identify an action or predict it from a partial sentence. The ConvNet architecture is simple and can be trained with minimum computational cost. We also demonstrate how important hyper-parameters such as data ratio are determined automatically. These parameters play significant roles in improving the accuracy. We achieve compatible state-of-the-art results on both action recognition and action prediction.

%{\small
%\bibliographystyle{ieee}
%\bibliography{bib}
%}
{\small
	\bibliographystyle{ieee}
	\bibliography{bib}
}

\end{document}